\documentclass{article}
\usepackage{spconf,amsmath,graphicx}
\usepackage{enumitem}
\setlist{nosep, leftmargin=14pt}
\usepackage{tabularx}
\usepackage{svg}
\usepackage{hyperref}
\usepackage{mwe} 

\title{FESS Loss: Feature-Enhanced Spatial Segmentation Loss for Optimizing Medical Image Analysis}
\name{Charulkumar Chodvadiya$^{ 2 , \ast}$ \qquad Navyansh Mahla$^{1, \ast}$ \qquad Kinshuk Gaurav Singh $^{2}$ \qquad Kshitij Sharad Jadhav$^{1}$ \thanks{$\ast$ equal contributors}} 
\address{$^{1}$ Indian Institute of Technology - Bombay, Mumbai, Maharashtra, India \\
$^{2}$ Pandit Deendayal Energy University, Gandhinagar, Gujarat, India}

\begin{document}

\maketitle
\begin{abstract} \label{Sec:abstract}
Medical image segmentation is a critical process in the field of medical imaging, playing a pivotal role in diagnosis, treatment, and research. It involves partitioning of an image into multiple regions, representing distinct anatomical or pathological structures. Conventional methods often grapple with the challenge of balancing spatial precision and comprehensive feature representation due to their reliance on traditional loss functions. To overcome this, we propose \textbf{Feature-Enhanced Spatial Segmentation Loss (FESS Loss)}, that integrates the benefits of contrastive learning (which extracts intricate features, particularly in the nuanced domain of medical imaging) with the spatial accuracy inherent in the Dice loss. The objective is to augment both spatial precision and feature-based representation in the segmentation of medical images. \textbf{FESS Loss} signifies a notable advancement, offering a more accurate and refined segmentation process, ultimately contributing to heightened precision in the analysis of medical images. Further, \textbf{FESS loss} demonstrates superior performance in limited annotated data availability scenarios often present in the medical domain. \footnote{\tiny \url{https://github.com/charulp2499/FeSS_Loss}}
\end{abstract}

\begin{keywords}
Medical Image Segmentation, Supervised Contrastive Learning, Feature Representation
\end{keywords}

\section{Introduction}\label{Sec:intro}
Advanced medical imaging technologies play a critical role in improving diagnostic precision, treatment and prognosis. Imaging modalities such as Magnetic Resonance Imaging (MRI), Computed Tomography (CT) etc are part of this sophisticated arsenal, and they provide nuanced insights into the complexity of human anatomy and pathology. Extracting clinically relevant information from these modalities demands computational finesse and medical image segmentation is at its core, utilizing both traditional procedures and cutting-edge deep learning techniques such as convolutional neural networks and U-Net architectures \cite{12,16,199}. Effectiveness of these methodologies relies on precise formulation of the loss function that measures the difference between expected and true segmentation maps. This computational sophistication, in conjunction with a meticulously designed loss function, enables accurate delineation of regions of interest\cite{19}.

\hspace{0.4cm}

\noindent\textbf{\large Our Contribution:} Contrastive learning is a method that distinguishes between positive and negative samples, improving feature extraction and classification accuracy \cite{20,22}. It is beneficial for medical imaging since it provides a detailed understanding of anatomical structures, allowing for precise diagnosis and segmentation of minute differences, while Dice loss, which serves as a metric for evaluating segmentation similarity, focuses on spatial overlap between predicted and ground truth segments \cite{19, 03}. Integrating Dice Loss into the contrastive learning framework produces a promising synthesis for segmentation problems, combining spatial accuracy with feature-based enhancement \textcolor{blue}{\textit{(c.f. sec. \ref{Sec:approach})}}. Combining these two improves feature representation, leading to better segmented medical image precision and characterization even in scenarios of limited annotated data  \textcolor{blue}{\textit{(c.f. sec. \ref{ssec:Varied_train_data})}}.

\section{Related work} \label{Sec:related_work}

Segmentation methods generate overlapping probability maps, however binary measures may be biased to structure size, resulting in artifacts or inaccuracies around object borders \cite{13,15}. Dice loss is applied for handling data imbalance especially between foreground and background. In situations where one class (usually the background) significantly outweighs the other in terms of the number of pixels or voxels,  Dice loss assigns higher importance to the smaller class (foreground) while training a segmentation model. Focal Dice Loss and Tversky index reduces imbalance by using balanced sampling during training and emphasized hard sample. Stochastic Aggregated Dice Coefficient (SA Dice) loss, a modified U-Net structure is used where batch segmentation outputs are merged into a larger image for dice loss calculation \cite{12,40,16,14}. simCLR and infoNCE utilized effective contrastive learning, by combining a learnable transformation and shows the benefits of larger batch sizes \cite{66,67}. In medical image analysis, a contrastive distillation paradigm enhances voxel-wise representation learning by predicting distance maps of object boundaries from dual views. It merges blocks in medical image segmentation based on intensity, position, and gradient features, maintaining boundary information but potentially resulting in fuzzy boundaries due to the lack of correlation feature calculation \cite{01, 09}.

\section{Our approach}\label{Sec:approach}
Our approach deliberately employs both spatial and feature-based information. Implementing Dice loss assures excellent spatial precision, while incorporation of contrastive loss improves feature representation quality, facilitating a robust segmentation process. We provide control over the sensitive trade-off between emphasizing correct segmentation (as characterized by the Dice loss) and boosting discriminative feature learning (as captured by the contrastive loss) by utilizing a hyperparameter \(\lambda\). This fine-grained adaptability translates into a refined optimization process in which our modified loss function capitalizes on the inherent capabilities of both loss components contributing significantly to a better generalization (as demonstrated through our experiments on the combined BraTs Dataset (Table \ref{tab:comp})) and improved performance. The equilibrium achieved through this combination is expressed in equation \ref{Loss_Our}.

\begin{equation}
\label{Loss_Our}
\text{FESS}_{\text{Loss}} = \lambda \times \text{Loss}_{\text{Dice}} + (1 - \lambda) \times \text{Loss}_{\text{Contrastive}}
\end{equation}

\noindent Here, $\lambda$ is a hyperparameter that determines the contribution of the Dice loss to the FESS loss, while the complement \((1 - \lambda)\) governs the influence of the contrastive loss.

\subsubsection{The Dice Loss}\label{Sec:dice_loss} 
Dice loss evaluates the overlap between two binary vectors' pixel-wise intersections, demonstrating the difference between the ground truth and predicted binary masks \cite{03,19}.

\begin{equation}
\label{dic_eq}
\footnotesize{
\text{Loss}_{\text{Dice}} = 1 - \frac{2 \sum_{i,j,k} y_{ijk} \cdot \hat{y}_{ijk} + \epsilon}{\sum_{i,j,k} y_{ijk} + \sum_{i,j,k} \hat{y}_{ijk} + \epsilon}    } 
\end{equation}

\noindent Here, \(y_{ijk}\) denotes the ground truth binary mask, and \(\hat{y}_{ijk}\) represents the predicted binary mask. The numerator quantifies twice the intersection between the masks, with \(\epsilon\) added for smoothness. The denominator calculates the sum of individual areas for both ground truth and predicted masks, added smoothness \(\epsilon\) ensuring avoidance of division by zero.

\subsubsection{The Contrastive Loss}\label{Sec:contra_loss}
Let \(x\) and \(y\) represent multidimensional vectors for the present and previous sample batch embeddings, respectively. This will facilitate the comparison of consecutive batch representations, enables consistent feature representation, and improve generalized learning. Every vector is defined by five indices, denoted as \(n, i, j, k, l\), signifying distinct dimensions within the embedding space. Here, \(n\) corresponds to the batch size, while \(i, j, k\) are indicative of spatial dimensions within the image, and \(l\) represents the feature index. The vectors are scaled along these dimensions by the inverse of their L2 norm for normalization to ensure adjustment to unit length. 
Since the vectors are placed on the unit hypersphere during this normalizing process, it is possible to compare the directional relationships between the vectors which is essential for similarity calculation. 
Also, the standardized representation is essential for robust similarity computations, allowing for extensive and reliable analysis of sample similarities. The computation of normalized similarity measure includes generating a total sum by element-wise multiplication of the normalized components of vectors \(x\) and \(y\) across dimensions \(i, j, k\) (Equation \ref{simi_eq}). Moreover, the use of a element wise multiplication between two normalized vectors maintains a keen sensitivity to variations in vector magnitudes which are crucial to our task. 

\begin{equation} \label{simi_eq}
\text{similarity} = \sum_{i,j,k} x_{ijk} \times y_{ijk}
\end{equation}

\noindent Here, \(x_{ijk}\) and \(y_{ijk}\) denote the normalized components of vectors \(x\) and \(y\) along the specified dimensions. The sum is taken over all dimensions, signifying the sum of the element-wise multiplication of the corresponding components of \(x\) and \(y\) and it captures the similarity by emphasizing the contribution of each dimension to the overall alignment between the two vectors. The subsequent steps in the contrastive loss calculation use this similarity measure to inform the model about the relationships between the sample embeddings. 

The numerator of the Contrastive loss function is given by \(\exp\left(\frac{\text{similarity}}{\Delta}\right)\), involves exponentiating the similarity between normalized vectors \(x\) and \(y\). The denominator, \(\sum_{i,j,k} \exp\left(\frac{\text{similarity}}{\Delta}\right)\), has a impact on how probabilities are calculated. For normalized vectors \(x\) and \(y\), it entails adding the exponentiated similarities across all dimensions \(i, j, k\). Each term in the sum corresponds to the exponentiated similarity value for a specific dimension. The $\Delta$ (Delta) is a temperature parameter that influences the spread and concentration of the resulting probabilities: a higher temperature results in a softer, more uniform distribution, while a lower temperature highlights disparities. Finally, we calculate \(\text{Loss}_{\text{Contrastive}}\) (equation \ref{con_eq}) by calculating the logarithm of the numerator-to-denominator ratio and scaling it by the learning rate \(\eta\) and dividing it by the number of batch samples \(N)\). 

\begin{equation}
\label{con_eq}
\small{
\text{Loss}_{\text{Contrastive}} =  \eta \times  ( -\frac{1}{N} \left[\log\left(\frac{\text{numerator}}{\text{denominator}}\right)\right])}
\end{equation}

The contrastive loss measures dissimilarity by taking into account the alignment of normalized vectors and penalizes deviations by using the negative log likelihood of softmax probabilities. The total optimization process is aided by the learning rate and batch samples adjustments.

\subsection{Integration With Model}

\begin{figure}[h]
  \centering
   \includegraphics[width=\linewidth]{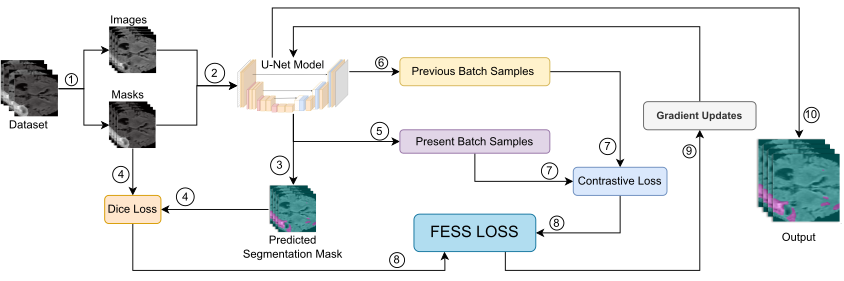}
    \caption{\footnotesize Illustrating Integration of FESS Loss into the 3D U-Net model and showing its calculation in the training loop with gradient updates}
  \label{FESS_FlowChart}
\end{figure}

We integrate our proposed Feature-Enhanced Spatial Segmentation Loss (FESS loss) with the U-Net 3D model providing a unique approach for reducing the issues of overfitting and weak predictions observed in the U-Net 3D model across diverse image datasets \cite{28}. This integration requires adjustments to both the contrastive and Dice losses, guiding the model to predict masks that not only match the ground truth but also integrate contextual information from previous batches. Specifically, the ground truth mask and the predicted mask are used to calculate the dice loss. Furthermore, as Flowchart \ref{FESS_FlowChart} illustrates, the calculation of the contrastive loss requires utilizing data from both the prior sample batch and the present sample batch. This comprehensive integration avoids model overfitting and improves segmentation accuracy capturing fine information in medical images and enhancing Model's generalization capabilities, allowing it to adapt well to the different spatial structures seen in medical images.

\begin{table*}[!h]
\centering
\footnotesize
\hspace{0.1cm}
\begin{tabularx}{\textwidth}{| p{2.5cm} | X | X | X | X | p{2.3cm} | X |} \hline
\textbf{Dataset} & \textbf{Loss Function} & \textbf{DICE cofficient} &  \textbf{IoU} & \textbf{Precision} & \textbf{Specificity} & \textbf{Sensitivity}\\
\hline
\hline

\textbf{BraTs 2016} & \textbf{Our Loss} & \textbf{0.85} $\pm$ \textbf{0.02} & \textbf{0.75} $\pm$ \textbf{0.03} & \textbf{0.80} $\pm$ \textbf{0.02} & 0.98 $\pm$ 0.001 & \textbf{0.91} $\pm$ \textbf{0.01}\\
\cline{2-7}
& \textbf{Dice Loss} & 0.69 $\pm$ 0.01 & 0.57 $\pm$ 0.03 & 0.70 $\pm$ 0.04 & 0.97 $\pm$ 0.003 & 0.72 $\pm$ 0.02 \\
\cline{2-7}
& \textbf{simCLR} & 0.70 $\pm$ 0.02 & 0.57 $\pm$ 0.01 & \textbf{0.80} $\pm$ \textbf{0.03} & \textbf{0.99} $\pm$ \textbf{0.001} & 0.67 $\pm$ 0.02\\
\cline{2-7}
& \textbf{infoNCE} & 0.71$\pm$ 0.02 & 0.58 $\pm$ 0.02 & 0.76 $\pm$ 0.03 & \textbf{0.99} $\pm$ \textbf{0.001} & 0.69 $\pm$ 0.02\\
\hline
\hline

\textbf{BraTs 2017} & \textbf{Our Loss} & \textbf{0.82} $\pm$ \textbf{0.03} & \textbf{0.71} $\pm$ \textbf{0.02} & 0.75 $\pm$ 0.04 & 0.97 $\pm$ 0.001 & \textbf{0.93} $\pm$ \textbf{0.01}\\
\cline{2-7}
& \textbf{Dice Loss} & 0.76 $\pm$ 0.02 & 0.63 $\pm$ 0.02 & 0.71 $\pm$ 0.04 & 0.97 $\pm$ 0.001 & 0.85 $\pm$ 0.02\\
\cline{2-7}
& \textbf{simCLR} & 0.77 $\pm$ 0.01 & 0.62 $\pm$ 0.03 & \textbf{0.80} $\pm$ \textbf{0.04} & \textbf{0.98} $\pm$ \textbf{0.002} & 0.77 $\pm$ 0.01 \\
\cline{2-7}
& \textbf{infoNCE}& 0.77 $\pm$ 0.02 & 0.64 $\pm$ 0.02 & \textbf{0.80} $\pm$ \textbf{0.03} & \textbf{0.98} $\pm$ \textbf{0.002} & 0.78 $\pm$ 0.04\\
\hline
\hline

\textbf{Combined BraTs} & \textbf{Our Loss} & \textbf{0.83} $\pm$ \textbf{0.03} & \textbf{0.69 }$\pm$ \textbf{0.03} & 0.86 $\pm$ 0.04 & \textbf{0.99} $\pm$ \textbf{0.001} & \textbf{0.77} $\pm$ \textbf{0.02} \\
\cline{2-7}
& \textbf{Dice Loss} & 0.73 $\pm$ 0.02 & 0.59 $\pm$ 0.02 & 0.76 $\pm$ 0.06 & \textbf{0.99} $\pm$ \textbf{0.002} & 0.74 $\pm$ 0.03 \\
\cline{2-7}
& \textbf{simCLR}& 0.71 $\pm$ 0.01 & 0.58 $\pm$ 0.02 & \textbf{0.94} $\pm$ \textbf{0.03} & \textbf{0.99} $\pm$ \textbf{0.002} & 0.62 $\pm$ 0.03\\
\cline{2-7}
& \textbf{infoNCE} & 0.81 $\pm$ 0.01 & 0.68 $\pm$ 0.01 & 0.92 $\pm$ 0.03 & \textbf{0.99} $\pm$ \textbf{0.001} & 0.74 $\pm$ 0.02 \\
\hline
\hline

\textbf{AbdomenCT-1K} & \textbf{Our Loss} & \textbf{0.63} $\pm$ \textbf{0.01}  & \textbf{0.46} $\pm$ \textbf{0.01} & \textbf{0.60}  $\pm$ \textbf{0.02}  & \textbf{0.96 } $\pm$ \textbf{0.003}  & 0.68 $\pm$ 0.04\\
\cline{2-7}
& \textbf{Dice Loss}& 0.60  $\pm$ 0.04 & 0.43  $\pm$ 0.04 & 0.55  $\pm$ 0.05 & 0.95  $\pm$ 0.004 & 0.67  $\pm$ 0.03\\
\cline{2-7}
& \textbf{simCLR} & 0.61  $\pm$ 0.01 & 0.44  $\pm$ 0.01 & \textbf{0.60}  $\pm$ \textbf{0.04}  & \textbf{0.96}  $\pm$ \textbf{0.004}  & 0.63  $\pm$ 0.04\\
\cline{2-7}
& \textbf{infoNCE} & 0.61  $\pm$ 0.01  & 0.44  $\pm$ 0.01 & 0.55  $\pm$ 0.03 & 0.95  $\pm$ 0.009  & \textbf{0.69}  $\pm$ \textbf{0.04} \\
\hline
    
\end{tabularx}

\caption{\small Quantitative Test Results for Segmentation Performance with FESS Loss and Different Functions on Medical Image Datasets}
\label{tab:comp}
\end{table*}

\section{EXPERIMENTS} \label{Sec:Experi}


\subsection{Datasets} \label{Sec: Datset}

We use three datasets in our experiments: the 2016 and 2017 \textbf{BraTs datasets} \cite{99} include high-quality multi-modal MRI scans (274 and 285 patients, respectively), including T1, T1-weighted post-contrast, T2, and FLAIR sequences. We use the FLAIR modality for our task. Sample diversity is improved by combining the datasets from BraTs 2016 and 2017. And we use \textbf{the AbdomenCT-1K dataset} \cite{26}, which includes over a thousand abdominal CT scans from various vendors and phases, is used to assess the generalizability of our approach. For our experiments, we use an 80:20 train:test data split from each dataset.

\subsection{Parameters Setting} \label{Sec:parameterSetting}
We used a batch size (\(N\)) of 5 throughout the test and training phases in order to get our optimal results. The optimum value of \({\text{1e}}-{\text{5}}\) was identified for obtaining appropriate model convergence in terms of the learning rate (\(\eta\)). Using a learning rate greater than \({\text{1e}}-{\text{5}}\) increased the possibility of overfitting or underfitting, which occurs when a model converges quickly. Additionally, the hyperparameters \(\epsilon\) (set at \({\text{1e}}-{\text{5}}\) to ensure smoothness in dice loss, and avoiding division by zero) and Temperature \(\Delta\) (set at 0.5 for contrastive loss to control probability spread and concentration) were utilized.

\section{Results \& Discussion} \label{Sec:Result}
\subsection{Evaluation of Results Across Diverse Datasets} \label{ssec:All_Result}
We endeavoured to improve on the existing loss functions used for medical image segmentation by incorporating Dice loss for
spatial precision, with contrastive loss to improves  feature representation quality, facilitating a robust segmentation. 
The evaluation of segmentation performance on multiple runs on different medical image datasets is quantified in Table \ref{tab:comp}. The results showcase the behavior of different loss functions, demonstrating the effectiveness of our FESS Loss in comparison. On the BraTs 2016 dataset, FESS Loss outperforms other loss functions by up to 16\%, with an average Dice coefficient of 0.85 indicating its effectiveness in capturing complex patterns in medical images. FESS Loss maintains its performance on the BraTs 2017 dataset, with a Dice coefficient of 0.82, outperforming other loss functions by approximately 7\%. While comparing FESS Loss to Dice Loss, simCLR, and infoNCE, the combined BraTs datasets show a notable 10\%, 12\%, and 3\% improvement, respectively, with a Dice coefficient of 0.83. Finally, to demonstrate that these improvements are not only on Brain MRI images, we tested FESS loss on the AbdomenCT-1K dataset and demonstrated a Dice coefficient of 0.63 and improvements of 5\%, 3\%, and 3\% over Dice Loss, simCLR, and infoNCE, respectively. These results indicate the flexibility of FESS Loss in processing a variety of medical imaging datasets and reliably generating very accurate segmentations.

\subsection{Qualitative Results}\label{ssec:Figure_Result}
\begin{figure}[!h]
  \centering
   
   \includegraphics[width=0.97\linewidth]{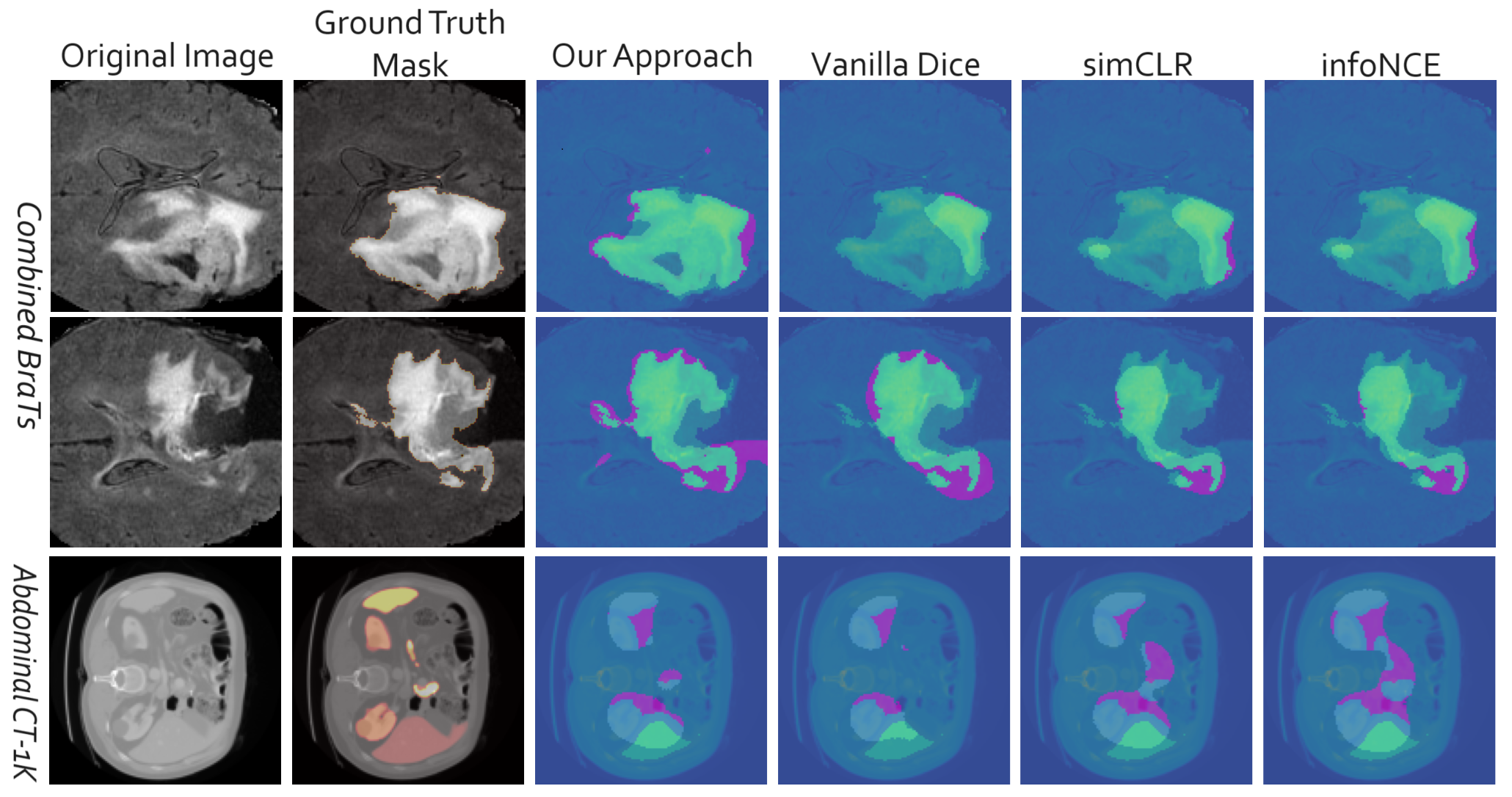}
    \caption{\footnotesize Comparison of segmentation results using our FESS Loss alongside baseline functions, showcasing the visual improved segmentation achieved by our framework}
  
  \label{Brain_Vis}
  \vspace{-10pt}
\end{figure}

The segmentation maps obtained from diverse datasets are depicted in Figure \ref{Brain_Vis}, which contrast Dice Loss, infoNCE, and simCLR to provide a unbiased assessment of our FESS loss on the abdominal organ and FLAIR modality for the brain. FESS loss produces segmentation masks that are close to the ground truth, even with varied baselines. FESS loss visually outperforms other baselines, leading to improved segmentation results.

\subsection{Evaluation Across Varied Training Data Sizes}\label{ssec:Varied_train_data}

\begin{figure}[!h]
  \centering
   \includegraphics[width=0.97\linewidth]{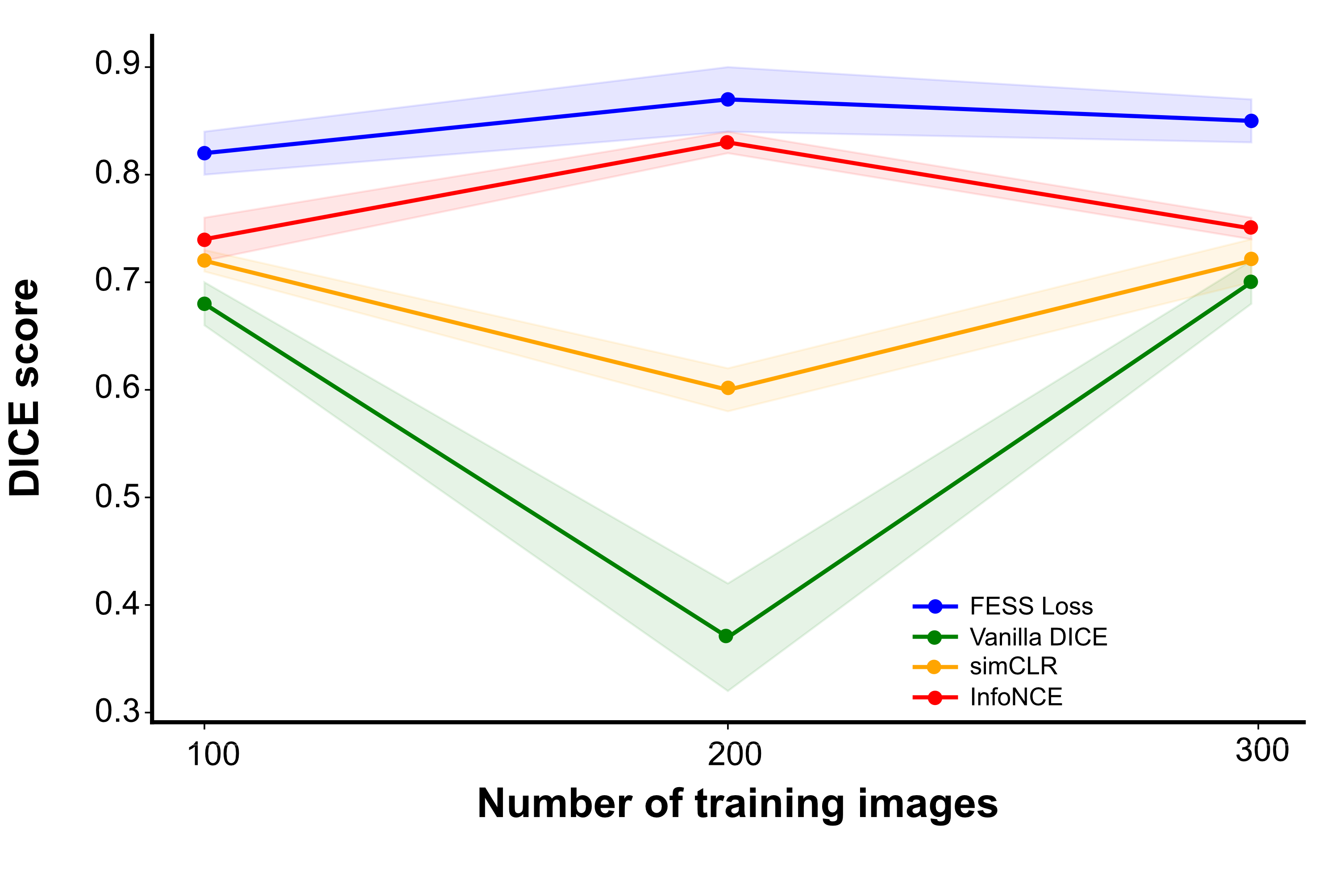}
    \caption{\footnotesize DICE score variation plot reveals insights into model performance stability, showing the mean and standard error across different sample sizes}
  \label{train_com_plot}
   \vspace{-20pt}
\end{figure}

To investigate the effectiveness of FESS Loss in limited data availability scenarios, we evaluated it by simulating small training data using the combined BraTs dataset. The outcomes, illustrated in figure \ref{train_com_plot}, highlight the superior performance of FESS Loss compared to baseline methods such as Dice Loss, simCLR, and infoNCE, especially in scenarios with a restricted number of training samples. Our Method, which began with a training set of 300 samples, achieved an impressive Dice coefficient of 0.82, surpassing the baseline scores of 0.68, 0.72, and 0.74. As the training set size decreased to 200 and then 100 samples, FESS Loss consistently demonstrated superior performance, with percentage improvements of up to 50\% and 15\%, respectively, compared to the other baselines. This underscores the efficacy of our approach in achieving heightened performance even with less training data. These results not only demonstrate the efficiency and effectiveness of FESS Loss in scenarios characterized by constrained data availability but also emphasize its potential as a promising solution for tasks with limited training resources, particularly in real-world test cases.

\section{Conclusion \& Future Work} \label{Sec:Concu._Future work}
Our proposed Feature-Enhanced Spatial Segmentation Loss (FESS Loss) offers an improved advancement in medical image segmentation. FESS Loss achieves a more precise and refined segmentation by combining the benefits of contrastive learning with the spatial accuracy provided by the Dice loss. The experimental results show improvements in the result matrix over a variety of medical image datasets, demonstrating its advantages over traditional loss functions. The primary strength of FESS Loss's its ability to balance feature representation with spatial accuracy, which helps address some of the drawbacks of traditional segmentation techniques. The integration of contrastive learning improves the extraction of intricate features in the complex domain of medical imaging, leading to a more sophisticated understanding of anatomical structures. This, coupled with the spatial precision ensured by the Dice loss, results in a comprehensive and refined segmentation process. Importantly, we demonstrate improved results with limited data which is a hallmark in medical imaging due to less data availability. Looking ahead, future work could explore additional refinements and applications of FESS Loss, further advancing its capabilities in the evolving landscape of medical image analysis.

\section{Compliance with ethical standards} \label{sec:ethics}
There are no conflict of  interest. Datasets are publicly accessible, they are all licensed and properly cited.


\bibliographystyle{IEEEbib}
\bibliography{refs}

\end{document}